\documentclass[10pt,twocolumn,letterpaper]{article}

\usepackage{cvpr}              

\usepackage[dvipsnames]{xcolor}

\definecolor{cvprblue}{rgb}{0.21,0.49,0.74}
\usepackage[pagebackref,breaklinks,colorlinks,citecolor=cvprblue]{hyperref}

\def\paperID{*****} 
\def\confName{CVPR}
\def\confYear{2024}

\title{vFusedSeg3D: 3rd Place Solution for 2024 Waymo Open Dataset Challenge in Semantic Segmentation}

\author{Osama Amjad\\
VisioRD\\
Pakistan\\
{\tt\small osama.amjad@visionrdai.com}
\and
Ammad Nadeem\\
VisionRD\\
Pakistan\\
{\tt\small ammad@visionrdai.com}
}

\begin{document}
\maketitle
\begin{abstract}
In this technical study, we introduce VFusedSeg3D, an innovative multi-modal fusion system created by the VisionRD team that combines camera and LiDAR data to significantly enhance the accuracy of 3D perception. VFusedSeg3D uses the rich semantic content of the camera pictures and the accurate depth sensing of LiDAR to generate a strong and comprehensive environmental understanding, addressing the constraints inherent in each modality. Through a carefully thought-out network architecture that aligns and merges these information at different stages, our novel feature fusion technique combines geometric features from LiDAR point clouds with semantic features from camera images. With the use of multimodality techniques, performance has significantly improved, yielding a state-of-the-art mIoU of 72.46\% on the validation set as opposed to the prior 70.51\%.VFusedSeg3D sets a new benchmark in 3D segmentation accuracy, making it an ideal solution for applications requiring precise environmental perception.
\end{abstract}    

\section{Introduction}

vFusedSeg3D is a multi-model architecture that used both lidar point cloud and camera images to fuse features and output enhanced feature map that can be used to predict segmentation maps. Our work is inspired by \cite{mseg3d_cvpr2023}, that also uses multimudalities to fuse camera and lidar features.

\section{Architecture}

\begin{figure}[h]
\centering
\fbox{\includegraphics[width=0.9\linewidth]{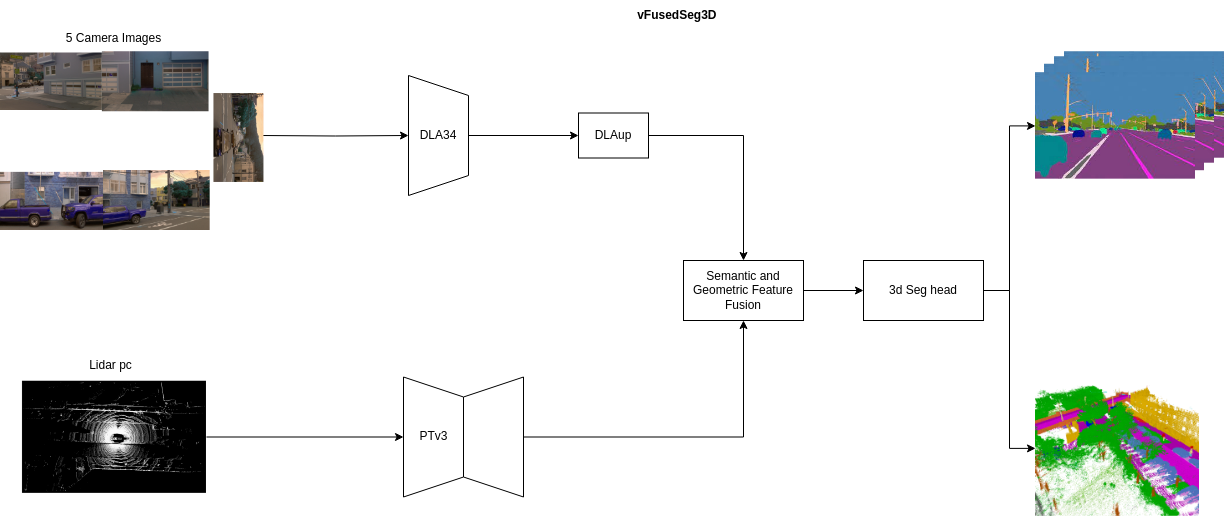}}
\caption{vFusedSeg3d architecture}
\label{fig:imageWithBorder}
\end{figure}

Our architecture uses dual-modal feature extraction to take advantage of the capabilities of both image and LiDAR data, resulting in robust feature representation.

The architecture consists of two parallel feature extraction streams. We utilized  DLA34, \cite{yu2018deep} as the image side backbone for feature extraction due to its hierarchical feature aggregation and multiple levels of residual blocks for multi scale feature extraction. The Waymo Open Dataset sample includes a point cloud with 64 beams and 5 RGB camera images from the following viewpoints: front, front-left, front-right, side-left, and side-right. These 5 images are utilized in their entirety. After resizing for model input, the images have a resolution of 960x640 (width x height).

The image encoder generates feature maps with dimensions of 16, 32, 64, 128, 256, and 512 feature channels. To facilitate detailed image feature extraction, a simple DLAup (Deep Layer Aggregation upsampling) module is employed as the neck of the image-side feature extraction. This module leverages residual connections to enhance the feature maps. This map is critical for subsequent fusion operations because this enable comprehensive scene knowledge by combining image-based information.

We used Point Transform v3 (PTv3), \cite{wu2024ptv3} as the key feature extraction backbone on the LiDAR side. PTv3 is useful in processing 3D point cloud data, extracting spatial features that encapsulate the geometric complexities of the environment.

\begin{figure}[h]
\centering
\fbox{\includegraphics[width=0.7\linewidth]{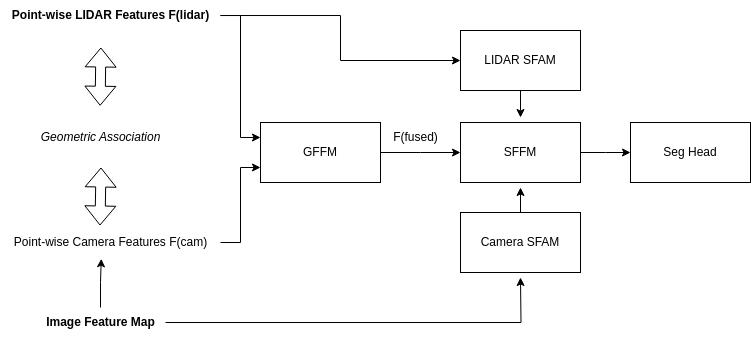}}
\caption{Lidar and camera feature fusion: Geometric Feature Fusion Module and Semantic Feature Fusion Modules as designed by \cite{mseg3d_cvpr2023} but with some modifications}
\label{fig:imageWithBorder}
\end{figure}

Following feature extraction as shown in , both image and LiDAR feature maps go through a fusion process to combine semantic and geometric data. The fusion process is based on a modified version of MSeg3D's feature fusion module that was designed specifically for this architecture. Point-wise LIDAR features and point-wise camera features are concatenated together subsequent learnable
fusion using a Geometry-based Feature Fusion Module (GFFM) resulting out geometric fused features. Lidar and Camera features are aggregated using there respectiv Semantic Feature Aggregation Module (SFAM). Instead of voxel Features as used by MSeg3d, we utilized point-wise lidar features to be aggregated using LIDAR SFAM. These aggregated features are used as the input of the the Semantic Feature Fusion Module (SFFM), which has been significantly improved, with new layers added to optimise the integration of semantic cues across modalities. This modified fusion technique allows for a more precise and detailed interpretation of the combined data, resulting in better performance in the following analytical tasks.

\begin{table}[ht]
\centering

\begin{tabular}{@{}p{3cm}p{5cm}@{}}
\toprule
\textbf{Parameter} & \textbf{Value} \\ \midrule
\textbf{Augmentations} & \begin{tabular}[c]{@{}p{5cm}@{}}
    \textbf{Lidar Side:} Global rotation around the Z-axis (\([- \pi/4, +\pi/4]\)), Global translation (\(\Delta x, \Delta y, \Delta z\) with \(N(0, 0.5)\)), Global scaling ([0.95, 1.05]). \\
    \textbf{Image Side:} Scaling ([1.0, 1.5]), Horizontal rotation ([\(-1^\circ, 1^\circ\)]), Random cropping (size \(H_{in}, W_{in}\)), Color jitter (brightness, contrast, saturation, hue with parameters 0.3, 0.3, 0.3, 0.1), JPEG compression ratio ([30, 70] with probability 0.5).
    \end{tabular} \\ \midrule
\textbf{Image mean/std} & mean = [0.40789654, 0.44719302, 0.47026115], std = [0.28863828, 0.27408164, 0.27809835] \\ \midrule
\textbf{Lidar enc Input Features} & 5 (x, y, z, intensity, elongation) \\ \midrule
\textbf{Point Cloud Serialization} & ["z", "z-trans", "hilbert", "hilbert-trans"] \\ \midrule
\textbf{Channels} & \begin{tabular}[c]{@{}p{9cm}@{}}
    \textbf{Lidar Enc:} (32, 64, 128, 256, 512) \\
    \textbf{Lidar Dec:} (64, 64, 128, 256) \\
    \textbf{Image Enc:} (32, 64, 128, 256, 512) \\
    \textbf{Image Neck:} (32, 64, 128, 256)
    \end{tabular} \\ \midrule
\textbf{Point Cloud Range} & (-75.2, -75.2, -4, 75.2, 75.2, 2) \\ \midrule
\textbf{Image Size} & (960, 640) (width, height) \\ \midrule
\textbf{Optimizer} & AdamW \\ \midrule
\textbf{Scheduler} & Cosine \\ \midrule
\textbf{Criteria} & CrossEntropy, Lovasz \\ \midrule
\textbf{Learning Rates} & \begin{tabular}[c]{@{}p{9cm}@{}}
    Main LR: 8e-4 \\
    Block LR: 8e-5
    \end{tabular} \\ \midrule
\textbf{Weight Decay} & 5e-2 \\ \midrule
\textbf{Batch Size} & 2 \\ \midrule
\textbf{Epochs} & 50 \\ \bottomrule
\end{tabular}
\centering
\caption{Training Hyperparameters for vFusedSeg3D}
\label{tab:hyperparameters}

\end{table}

\section{Training Strategy}

Due to computational resource constraints, we trained the model sequentially and in segments. We initially approached this difficulty by training the LiDAR and image feature extraction components separately. This method allowed us to focus resources effectively, ensuring that each component was optimized to its full potential without overwhelming our hardware capabilities.

For the LiDAR (lidarBase) side, we trained the PTv3 model for 45 epochs with a batch size of 2. The training process spanned approximately 6 to 7 days, culminating in a highest achievable accuracy of 70.51\% mIoU. In the original paper by \cite{wu2024ptv3}, the reported accuracy is around 71.3\% mIoU.

The observed decrease in accuracy is primarily attributable to differences in grid sampling resolution and low batch size. The original study employed a finer grid sampling of 0.05 and batch size of 12 with multi-GPU training, whereas we used a coarser grid sampling of 0.1, due to limitations in our computational resources, and batch size of 2 with single GPU. Increasing the grid resolution beyond this value resulted in Out of Memory (OOM) errors, thus constraining our ability to achieve the higher accuracy reported in the original paper.Image side was trained for 10 epochs with batch size of 8.

Once the individual LIDAR and image models were trained and their parameters fine-tuned, we froze these models to conserve resources. This freezing process prevented any further modifications to the weights of the feature extraction models during the subsequent training phases, thereby reducing the computational load. This fusion model was trained for 25 epochs with batch size of 2. We did not use mini-batching for gradient accumulation, although it is a good idea, but due to time limitation for more experimentation, we trained our models on 2 batch size.

Subsequently, we directed our resources towards training the fusion model, which integrates the pre-trained and frozen image and LiDAR feature maps. This staged training approach not only addressed our resource limitations but also ensured that the integration layer—responsible for merging semantic and geometric features—received the dedicated attention needed for effective learning.

By compartmentalizing the training process in this manner, we managed to circumvent the limitations imposed by our hardware resources, allowing each part of the network to be meticulously trained and thereby enhancing the overall performance and efficiency of the model. This structured training strategy proved crucial in developing a robust multi-modal fusion system capable of high-accuracy performance in real-world applications.

\begin{table}[ht]

\begin{tabular}{@{}lc@{}}
\toprule
\textbf{Model} & \textbf{mIoU (\%)} \\
\midrule
Our LidarBase  & 70.51 \\
PTv3 \cite{wu2024ptv3} & 71.3 \\
vFusedSeg3D & 72.46 \\
\bottomrule
\end{tabular}
\centering
\caption{Model Performance and mIoU}
\label{tab:model_performance}
\end{table}

\section{Final Results on VAL set}

As demonstrated in Table~\ref{tab:model_performance}, the model performance metrics are summarized based on their mean Intersection over Union (mIoU) percentages.

\section{Training Hyper parameters}

Training parameters used for model training are given in the Table~\ref{tab:hyperparameters}.

\section{Test Time Augmentations (TTA)}

We only utilized Test Time Augmentations on lidar point clouds to boost our accuracies as mentioned on Table~\ref{tab:test_time_augmentations}

\begin{table}[ht]

\begin{tabular}{@{}lp{5cm}@{}}
\toprule
\textbf{Augmentation} & \textbf{Description} \\ \midrule
Global Scaling (\(\tau_{\text{scale}}\)) & A random scaling factor in [0.95, 1.05]. \\ 
Random Flipping (\(\tau_{\text{flip}}\)) & Random flipping along the X, Y axis. \\ 
Global Rotation (\(\tau_{\text{rot}}\)) & Rotation around the Z axis with a random angle in \([- \frac{\pi}{4}, +\frac{\pi}{4}]\). \\ 
Global Translation (\(\tau_{\text{tran}}\)) & Translation with a random vector (\(\Delta x, \Delta y, \Delta z\)). \\ \bottomrule
\end{tabular}
\centering
\caption{Test Time Augmentations (TTAs)}
\label{tab:test_time_augmentations}
\end{table}

\section{Experimental Setup}

All experiments and training sessions were conducted on a single NVIDIA RTX 3090 graphics card, equipped with 24GB of VRAM. The GPU was paired with 32GB of CPU RAM.

{
    \small
    \bibliographystyle{ieeenat_fullname}
    \bibliography{main}

\begin{thebibliography}{3}
\providecommand{\natexlab}[1]{#1}
\providecommand{\url}[1]{\texttt{#1}}
\expandafter\ifx\csname urlstyle\endcsname\relax
  \providecommand{\doi}[1]{doi: #1}\else
  \providecommand{\doi}{doi: \begingroup \urlstyle{rm}\Url}\fi

\bibitem[Li et~al.(2023)Li, Dai, Han, and Ding]{mseg3d_cvpr2023}
Jiale Li, Hang Dai, Hao Han, and Yong Ding.
\newblock Mseg3d: Multi-modal 3d semantic segmentation for autonomous driving.
\newblock In \emph{CVPR}, pages 21694--21704, 2023.

\bibitem[Wu et~al.(2024)Wu, Jiang, Wang, Liu, Liu, Qiao, Ouyang, He, and Zhao]{wu2024ptv3}
Xiaoyang Wu, Li Jiang, Peng-Shuai Wang, Zhijian Liu, Xihui Liu, Yu Qiao, Wanli Ouyang, Tong He, and Hengshuang Zhao.
\newblock Point transformer v3: Simpler, faster, stronger.
\newblock In \emph{CVPR}, 2024.

\bibitem[Yu et~al.(2018)Yu, Wang, Shelhamer, and Darrell]{yu2018deep}
Fisher Yu, Dequan Wang, Evan Shelhamer, and Trevor Darrell.
\newblock Deep layer aggregation.
\newblock In \emph{Proceedings of the IEEE conference on computer vision and pattern recognition}, pages 2403--2412, 2018.

\end{thebibliography}
}


\end{document}